# A Stable Minutia Descriptor based on Gabor Wavelet and Linear Discriminant Analysis

Gwang-Il Ri[1,*], Mun-Chol Kim[1], Su-Rim Ji[2]

[1] Faculty of Mathematics, **Kim Il Sung** University, Pyongyang, Democratic People's Republic of Korea

[2] Electronic Library, **Kim Il Sung** University, Pyongyang, Democratic People's Republic of Korea

[*] Corresponding author

gi.ri@ryongnamsan.edu.kp

The minutia descriptor which describes characteristics of minutia, plays a major role in fingerprint recognition. Typically, fingerprint recognition systems employ minutia descriptors to find potential correspondence between minutiae, and they use similarity between two minutia descriptors to calculate overall similarity between two fingerprint images. A good minutia descriptor can improve recognition accuracy of fingerprint recognition system and largely reduce comparing time. A good minutia descriptor should have high ability to distinguish between different minutiae and at the same time should be robust in difficult conditions including poor quality image and small size image. It also should be effective in computational cost of similarity among descriptors. In this paper, a robust minutia descriptor is constructed using Gabor wavelet and linear discriminant analysis. This minutia descriptor has high distinguishing ability, stability and simple comparing method. Experimental results on FVC2004 and FVC2006 databases show that the proposed minutia descriptor is very effective in fingerprint recognition.

Keywords: Fingerprint Recognition, Minutia Descriptor, Linear Discriminant Analysis, Gabor Wavelet

## 1. Introduction

Fingerprint recognition is one of the most widely used biometrics recognition techniques as it has very high recognition accuracy, it is very simple to use and fingerprint sensor's price is very cheap. Automated fingerprint recognition has the longest research history among biometrics recognition and a lot of algorithms have been proposed in the literature. But most of the fingerprint recognition systems suffer from poor quality images, small valid regions and plastic distortion. Thus many researchers challenge this problem. In Figure 1 and Figure 2, images of a small size fingerprint, low quality fingerprints and distorted fingerprints are shown.

A fingerprint consists of ridges and valleys. In the fingerprint images ridge part is dark, while valley part is bright. Features for fingerprint recognition include global features and local features. Type of fingerprint, ridge orientation and ridge density information belong to the global features. Minutia information is typical of local features. Minutia refers to the point where the ridge flow has discontinuity property. There are various types of minutiae, but typical fingerprint recognition algorithms use only two types of minutiae; termination and bifurcation of ridges. Figure 3 shows minutiae of two types.

Generally, fingerprint recognition methods are grouped into three classes; correlation-based method, ridge-based method and minutia-based method [8]. Correlation-based methods compare two fingerprint images by calculating global correlation or local correlation using displacement and rotation.

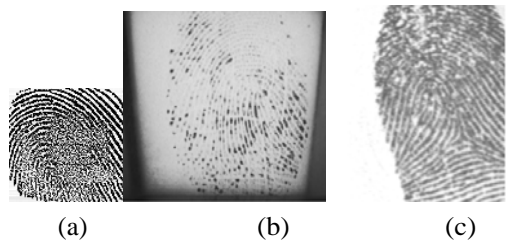

(a)  (b)  (c)

Fig. 1. Examples of a small size fingerprint image and poor quality fingerprint images.
(a) a small size image. (b), (c) poor quality images.

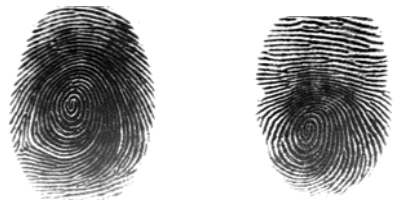

Fig. 2. Plastic distortion of fingerprints

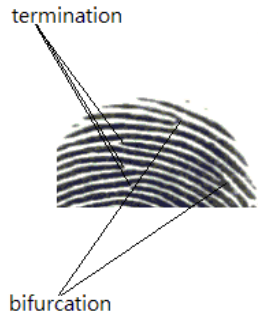

Fig.3. Two types of minutiae

Ridge-based methods compare two fingerprint images by using local orientation and frequency information of ridges, ridge shape and texture information of the images. Minutia-based methods compare two fingerprint images by extracting minutiae from images and matching two minutiae sets.

Minutia-based methods are most robust and they have the highest recognition accuracy among the fingerprint recognition methods. Recently, researches on minutia-based method to improve the recognition performance have been conducted including research for increasing stability of minutia extraction, research for improving the matching algorithm and research for constructing good feature information, etc. As good feature information allows the system to increase recognition accuracy without much additional computational cost, it is a very important step to extract good feature information for high performance of the system. Most of the minutia-based fingerprint recognition systems obtain feature information which describes the characteristics of every minutia of fingerprint and use it for matching. Positional and directional information of minutia, which are traditional index for describing minutia, are blind to distinguish two different minutiae. Thus, if only position coordinates and directional feature are used for comparison, every minutia of the test fingerprint should be compared with every minutia of the enrollment fingerprint. Adding more feature information which reflects characteristics of every minutia into positional and directional information makes correspondence probability between minutiae distinguishable each other. This feature information constructed for every minutia to distinguish different minutiae is called a minutia descriptor.

In the paper [1], the authors divided minutia descriptors into three classes: image-based, texture-based and minutia-based descriptors. Image-based descriptors employ raw image information around minutia as a feature for the minutia. The raw image information is acquired in the surrounding area centered at the minutia position and oriented towards the minutia direction. Texture-based descriptors compose feature vectors using orientation and frequency information of ridges. For example, in the paper [9] the authors proposed an orientation-based minutia descriptor using ridge orientations of sampling points surrounding minutia. In the paper [2], the authors proposed a texture-based minutia descriptor using ridge orientation and frequency information of sampling points surrounding minutia. And minutia-based descriptors are composed using geometrical relation between minutia and neighboring minutiae. Generally, minutia-based descriptors differ in the way how to define neighboring minutiae, i.e., the nearest neighbors of a certain number can be defined as the neighboring minutiae or all the minutiae within a fixed distance can be defined as the neighboring minutiae. In the paper [10], the authors proposed a minutia-based minutia descriptor using geometrical relations between $m$ neighboring minutiae and the central minutia. In the paper [4, 5], MCC is a kind of minutia-based descriptors composed using all neighboring minutiae in a fixed distance.

In the paper [1], the distinguishing ability of three types of minutia descriptors are compared on four fingerprint image conditions. As a conclusion, the texture-based descriptor shows the highest performance for small size images and MCC is the best for the rest three conditions including good quality, poor quality and plastic distortion. In the paper [2, 3], the authors combined various descriptors to further improve the distinguishing ability of minutiae and fingerprint recognition accuracy. In the paper [2], the authors assigned a texture-based descriptor and a minutia-based descriptor to each minutia and tested their performance on all four databases of FVC2002. In the paper [3], the authors combined minutia-based descriptor, Gabor transform-based descriptor and orientation information-based descriptor, and illustrated the effectiveness of the proposed method on the FVC2002 database.

In this paper we construct a novel minutia descriptor using Gabor wavelet and linear discriminant analysis, and demonstrate that the proposed descriptor has high distinguishing ability and stability in the various conditions, and it is very simple to compare.

This paper is organized as follows. In Section 2, we explain how to construct a minutia descriptor using Gabor wavelet and Linear Discriminant Analysis. In Section 3, we evaluate the effectiveness of the proposed minutia descriptor. Finally in Section 4, conclusions are made.

## 2. Constructing a Minutia Descriptor Using Gabor Wavelet and Linear Discriminant Analysis

Typical minutia-based matching methods consist of local matching and global matching. In the local matching step, similarity between every minutia of the enrollment template and every minutia of the test template is computed. Based on the similarity some reference minutia pairs with high correspondence probability are selected. In the global matching step, the test template is transformed into a coordinate system of the enrollment template with respect to every reference minutia pair with high correspondence probability which is selected in the local matching step, and minutia matching pairs are found using geometrical correspondence (position coordinates, directional information, etc.) and local similarity. Then the overall similarity is computed using the minutia correspondence and local similarity of the minutia matching pair. For every reference minutia pair selected in the local matching step, the above overall similarity is calculated and finally the maximum value is computed. For fingerprint recognition algorithms of this kind, the minutia descriptor can be used as follows.

   a) To find good minutia correspondence pairs in the local matching step

   Some minutia pairs with high correspondence probability are selected in the local matching step using similarity degree between minutia descriptors of two minutiae in minutia pair.

   b) To determine minutia correspondence pairs in the global matching step

   If the similarity degree between minutia descriptors of two minutiae in a minutia pair is lower than a certain threshold, then we decide that there is no correspondence between the two minutiae.

   c) To compute the overall similarity in the global matching step

   For computation of the overall similarity, not only the numbers of the correspondence minutiae, but similarity degree between minutia descriptors of the correspondence minutia are considered.

From this view, the minutia descriptor should have the following properties.

   i. Operation complexity for comparing between minutia descriptors should be low.

   In the local matching step, similarity degree between minutia descriptors from all the possible minutia pairs is computed. Thus, operation complexity for comparing should be low. Otherwise it will consume much time in overall matching.

   ii. Distinguishing ability should be high.

   The higher the descriptor's distinguishing ability is, the less number of reference minutia pairs the local matching step finds, therefore the less time the global matching step consumes. And also the higher it is, the more ability of removing false correspondence pairs has and the higher matching accuracy for overall similarity is.

   iii. It should be stable on various difficult conditions.

   Here, "stable" means that the minutia descriptor constructed for the same minutia is always steady irrespective of image quality and size. Commonly, for fingerprints of sufficient size and high quality, fingerprint recognition algorithms based on minutia show high accuracy. But for fingerprints of small size and poor quality, the performance of the recognition algorithms largely differs. That is why we need to construct a stable minutia descriptor to improve the performance of the overall fingerprint recognition.

   iv. New feature, which is independent of the information used already for matching fingerprint images, should be included in it.

   If and only if the minutia descriptor contains the information which is independent of the information used already, then the performance of the fingerprint recognition algorithm based on it would be improved significantly. Minutia-based descriptor is constructed using geometrical relation between the central minutia and neighboring minutiae. If the fingerprint image is of small size or poor quality, therefore if the accuracy of the preprocessing step is poor, then the neighboring minutiae set changes and the descriptor based on it would not be stable. Furthermore, most of the minutia-based algorithms have already employed the geometrical relation between minutiae in various ways; thus, there would not be much strong independence between information used already and minutia-based descriptor. Meanwhile the image-based minutia descriptor is not so sensitive to image size and accuracy of preprocessing as it depends only on positional and directional information of the minutia. But since it uses raw image as feature information, it is vulnerable to poor image quality and plastic distortion, and its discriminant ability is not so high. Texture-based minutia descriptor which uses ridge orientation and frequency information of sampling points around minutia has certain discriminant ability, but its stability mainly depends on stability of estimating ridge orientation and frequency from all the sampling points around the minutia. But poor quality images yield unreliable estimation.

In this paper, we propose a stable minutia descriptor based on Gabor wavelet and linear discriminant analysis. The proposed descriptor has high discriminant ability and low operational complexity as it only uses positional and directional information of the central minutia and the image information around it, and it does not employ any information of the neighboring minutiae, estimated ridge orientation or frequency of sampling points around the minutia. In the next subsection we explain how to construct our minutia descriptor and how to utilize it for fingerprint recognition.

### 2.1 Constructing a Minutia Descriptor

First, we calculate Gabor wavelet coefficients for every minutia and fixed sampling points around it. And next, we get fixed-length vectors by linear discriminant analysis. This fixed-length vector is just our minutia descriptor. Below, let us describe the procedure in detail.

① Extracting minutiae from fingerprint images

We extract minutiae from a fingerprint image using the minutia extraction algorithm proposed in paper [7].

② Fingerprint image enhancement

First, we do DOG filtering on the original fingerprint image.

Here, the impulse response of DOG filter is as follows:

$$D(x,y) = \frac{1}{2\pi\sigma_1}\exp\left(-\frac{x^2+y^2}{2\sigma_1^2}\right) - \frac{1}{2\pi\sigma_2}\exp\left(-\frac{x^2+y^2}{2\sigma_2^2}\right), \quad \sigma_1 < \sigma_2$$

The values $\sigma_1$ and $\sigma_2$ are set according to the resolution of the fingerprint image. Next, we do local mean variance regularization for the DOG filtered image.

$$I'(x,y) = \frac{c \times (I(x,y) - m)}{var^{\frac{1}{2}}} + 128$$

In the above equation, $I(x,y)$ is the resulting image after DOG filtering, $I'(x,y)$ is the resulting image after local mean variance regularization, $m$ and $var$ are the mean and the variance of a local area centered at the pixel $(x,y)$, respectively and $c$ is a fixed constant. Figure 4 shows the result of the image enhancement.

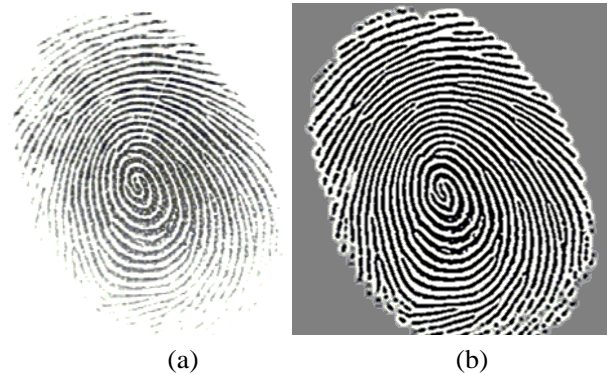

(a)        (b)

Fig. 4. (a) Original fingerprint image, (b) Resulting image after DOG filtering and local mean variance regularization.

③ Calculating Gabor wavelet coefficients for every minutia

We take 9 sampling points with one minutia as the center and the minutia direction as the reference direction as shown in Figure 5.

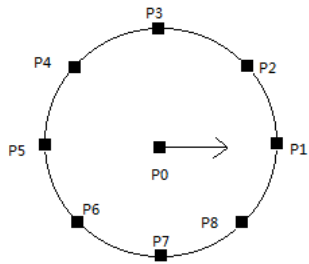

Fig. 5. 9 sampling points centered at one minutia

In the Figure 5, P0 is the position of the central minutia and the X-axis is the direction of the minutia. In this paper we take 18 for the radius. We calculate the absolute value of Gabor wavelet coefficients with 5 frequencies and 8 directions for all the 9 sampling points using the above-mentioned enhanced image. The 2-D Gabor wavelet function which is used for wavelet coefficient is as follows.

$$g_{\mu,\nu}(z) = \frac{\|k_{\mu,\nu}\|^2}{\sigma^2} \exp\left(-\frac{\|k_{\mu,\nu}\|^2 \|z\|^2}{2\sigma^2}\right)\left(\exp(ik_{\mu,\nu}z) - \exp\left(-\frac{\sigma^2}{2}\right)\right)$$

Here, $\mu$ and $\nu$ are the directional index and the frequency index of Gabor wavelet, respectively. $z = (x, y)$ is the pixel's position vector and $k_{\mu,\nu} = k_\nu(\cos\varphi_\mu, \sin\varphi_\mu)$, where

$$k_\nu = \frac{k_{\max}}{2^{\nu/2}}, \nu = 0,1,2,3,4$$

$$\varphi_\mu = \frac{\pi\mu}{8}, \mu = 0,1,2,...7$$

For every sampling point we calculate 40 absolute values of Gabor wavelet coefficients, concatenate them and assign an acquired 360-D vector (=40×9) to every minutia.

④ Construction of linear discriminant analysis transform

For linear discriminant analysis we made database ourselves using optical fingerprint sensor. The database consists of images of 500 fingers and 10 images are captured for every finger. For correct training, fingerprint images of poor quality are removed from the database. For every finger, the first image is enrolled and the rest 9 images are used for matching. For feature extraction and matching we employ the algorithm proposed in paper [7]. We do matching and at the same time, improve the template by the method proposed in paper [6]. But unlike the method in the paper [6], we store the 360-D Gabor wavelet coefficients vector of all the minutiae participants for improvement as well as the position and direction of the improved minutia as information of every minutia of enrollment template. As the result of matching step of 9 fingerprint images for every finger, at least one or at most ten 360-D vectors are made for every minutia of improved template. After matching step, we consider every minutia from the improved enrollment templates as one class and implement principal component analysis and linear discriminant analysis in the 360-D Gabor wavelet feature space. In other words, first we get a linear transform by principal component analysis to reduce 360-D space into 30-D one, next, we get a linear discriminant transform from 30-D to 25-D by linear discriminant analysis in the 30-D feature space and last, we get a 360×25 final linear transform multiplying the principal component transform (360×30) by the linear discriminant transform (30×25).

⑤ Converting the 360-D Gabor wavelet coefficients vector into a 25-D vector by linear discriminant transform

We multiply the transpose of the 360×25 linear discriminant transform by the 360-D Gabor wavelet coefficient vector to get a minutia descriptor for every minutia. After all, we get a 25-D fixed-length vector as a minutia descriptor for every minutia.

2.2. Exploiting the Minutia Descriptor

In this paper, the similarity between descriptors of two minutiae is calculated as follows. Let $V1, V2$ denote the descriptors of two minutiae, respectively. The similarity between two descriptors is calculated as follows.

$$SimD(V1, V2) = \log\left(\frac{\alpha}{\beta + Ed(V1, V2)}\right)$$

Here, $Ed(V1,V2)$ is the Euclid distance between $V1$ and $V2$. $\alpha$ is a large certain constant, $\beta$ is a very small certain constant. As you see, this similarity emphasizes the minutia pair which has small Euclid distance between two minutiae. We use two-step matching for fingerprint matching. In the first step, the elastic minutia comparison-based fingerprint matching proposed in paper [7] is employed. To do this, first, reference minutia pairs with high correspondence probability are selected. Next, for every reference minutia pair, we transform the coordinates according to the pair and perform minutiae matching using the algorithm of paper [7] to get the similarity. Among all the reference minutia pairs the maximum similarity pair is selected and all the correspondence minutia pairs are stored. Let $Sim1$ denote the maximum similarity value. In the second matching step, we get the similarity which reflects the degree of geometrical consistency between the minutiae pairs found in the first matching step. Let $Sim2$ denote this similarity. The overall similarity is $Sim = Sim1 \times Sim2$.

We used the proposed minutia descriptor three times during matching procedure, that is, to find the reference minutia pair of highest correspondence probability in the local matching step, to remove false minutia correspondence pairs using the similarity of the minutia descriptor in the global matching step and to calculate the first similarity $Sim1$.

- Selecting the reference minutia pair of highest correspondence probability.

Let $N1$ and $N2$ denote the number of minutiae of enrollment template and the number of minutiae of test template, respectively. For all the possible pairs of enrollment minutia and test minutia, we calculate the similarity between minutia descriptors. Then the calculated similarities are sorted in decreasing order. First $\lfloor N1 \times N2 \times pro/100 \rfloor$ minutia pairs with high similarity are selected as reference minutia pairs. Here, $pro$ is a certain constant.

- Removing false minutia correspondence pairs

If the similarity between descriptors of minutia pair is smaller than a certain threshold, we decide that the minutia pair does not have correspondence.

- Calculating the first similarity $Sim1$

In the paper [7], the authors calculated the matching similarity between enrollment template and test template as follows.

$$Sim1 = \frac{100 \times N_{pair}}{\max\{N1, N2\}}$$

In this paper we calculate the first similarity $Sim1$ using the similarities between descriptors.

$$Sim1 = \frac{100 \times \sum_{i=1:N_{pair}} SimD_i}{\max\{N1, N2\}}$$

Here, $SimD_i$ denotes the similarity between minutia descriptors of the $i$-th correspondence pair.

3. Experimental Results

We conducted experiments to evaluate the improvement of matching accuracy and effectiveness in selecting reference minutia pair by using the proposed minutia descriptor. In the experiments we used FVC2004 DB1, DB2, FVC2006 DB2, DB4 as test database. In FVC2004 DB1, DB2 and FVC2006 DB2 the images are acquired by optical fingerprint sensor and in FVC2006 DB4 the images are formed by artificial fingerprint image generation algorithm. For three fingerprint recognition systems, we evaluated the performance and effectiveness of the proposed minutia descriptor in the above databases.

The first system is the baseline system, which composes the similarity acquired by elastic matching algorithm in paper [7] and the similarity acquired by confirming step to get overall similarity. In the baseline system we did not use minutia descriptor. For the second system, we added a condition that the similarity between minutia descriptors should be larger than a certain threshold to the baseline system. And in the second system, the similarity of descriptor was used for calculating matching similarity.

For the third system, in order to reduce matching time we added selecting reference minutia pairs by using the proposed minutia descriptor to the second system. Table 1 shows that the proposed minutia descriptor is highly effective in improvement of matching accuracy by comparing the matching accuracy of the first system and of the second system. We evaluate the accuracy of fingerprint recognition with two indicators, i.e. Equal Error Rate and False Rejection Rate at 0.01 % of False Acceptance Rate. Experiments show that the accuracy of the system with the minutia descriptor is much improved in comparison with the accuracy of the system without minutia descriptor. Especially, the performance improvement in FVC2006 DB2 and FVC2006 DB4 is very significant since FRR (FAR=0.01%) is as large as 3~4 times.

Table 1. EER and FRR (FAR=0.01%) of the System 1 and the System 2.

|  |  | System 1 | System 2 |
|---|---|---|---|
| FVC2004 DB1 | EER | 7.087 | 5.70 |
|  | FRR ( FAR=0.01% ) | 20.25 | 14.92 |
| FVC2004 DB2 | EER | 4.17 | 2.65 |
|  | FRR ( FAR=0.01% ) | 12.39 | 6.93 |
| FVC2006 DB2 | EER | 0.137 | 0.063 |
|  | FRR ( FAR=0.01% ) | 0.638 | 0.152 |
| FVC2006 DB4 | EER | 0.944 | 0.411 |
|  | FRR ( FAR=0.01% ) | 4.567 | 1.537 |

Table 2 shows the experimental results of selecting reference minutia pairs by using the proposed minutia descriptor. The table explains how the percentage of the numbers of selected reference minutia pairs in the product of minutia numbers of two templates affects the matching accuracy.

Table 2. Percentage of Reference Minutia Pair Using Minutia Descriptor vs. Accuracy

|  |  | 1% | 2% | 3% | 4% | 5% |
|---|---|---|---|---|---|---|
| FVC2004 DB1 | EER | 6.568 | 5.997 | 5.723 | 5.665 | 5.665 |
| | FRR (FAR=0.01%) | 17.036 | 15.036 | 14.857 | 14.857 | 14.821 |
| FVC2004 DB2 | EER | 2.805 | 2.749 | 2.700 | 2.680 | 2.670 |
| | FRR (FAR=0.01%) | 7.536 | 7.107 | 7.0 | 6.893 | 6.857 |
| FVC2006 DB2 | EER | 0.068 | 0.058 | 0.058 | 0.058 | 0.063 |
| | FRR (FAR=0.01%) | 0.119 | 0.152 | 0.141 | 0.141 | 0.141 |

As you can see from the table, selecting only 2~3 % of the product of minutia numbers of two templates is enough to improve the accuracy.

As a conclusion, the proposed minutia descriptor is effective in matching accuracy improvement and very useful to select reference minutia pairs with low computational cost.

4. Conclusion

In this paper, we proposed a novel minutia descriptor constructing method using Gabor wavelet and linear discriminant analysis. As the minutia descriptor is a vector of fixed length and the similarity between the descriptors is based on the Euclid distance, comparing between two descriptors is very simple. Besides, it is not so sensitive to the accuracy of image preprocessing result unlike the other minutia descriptors, since our minutia descriptor only uses positional information and directional information of the minutia and image information around it, not employing neighboring minutiae information, estimated ridge orientation or density of neighboring sampling points. Moreover, the minutia descriptor steady reflects information (orientation change, ridge density information) of the image surrounding the minutia, since it employs the absolute value of Gabor wavelet coefficients of surrounding sampling points. And we reduced the length of the minutia descriptor and raised the discriminant ability by using linear discriminant analysis. Experimental results show that the proposed minutia descriptor can reduce matching time significantly by selecting the reference minutia pairs effectively and it can also largely improve matching accuracy by reflecting the similarity between minutia descriptors.